%% file: main.tex
\documentclass[letterpaper, 10 pt, conference]{ieeeconf}  

\IEEEoverridecommandlockouts                               

\usepackage[table, dvipsnames]{xcolor}
\usepackage{amsmath} 
\usepackage{graphicx}
\usepackage{adjustbox}
\usepackage{multicol}
\usepackage{tabularx}
\usepackage{caption}
\usepackage{amssymb}
\usepackage{cellspace}
\usepackage{hyperref}
\usepackage{cite}
\usepackage{multirow}
\usepackage{float}
\usepackage{booktabs}

\title{\LARGE \bf
DG16M: A Large-Scale Dataset for Dual-Arm Grasping with Force-Optimized Grasps}

\author{Md Faizal Karim$^{1*}$, Mohammed Saad Hashmi$^{1*}$, Shreya Bollimuntha$^{1}$, Mahesh Reddy Tapeti$^{1}$, Gaurav Singh$^{1}$, \\ Nagamanikandan Govindan$^{2}$, K Madhava Krishna$^{1}$ \vspace{1.5mm}\\
*Equal Contribution, $^{1}$Robotics Research Center, IIIT Hyderabad, $^{2}$ IIITDM Kancheepuram
}

\begin{document}

\thispagestyle{empty}
\pagestyle{empty}


\twocolumn[{
\renewcommand\twocolumn[1][]{#1}
\maketitle
\begin{center}
\vspace{-5mm}
    \centering
    \captionsetup{type=figure, font = footnotesize} 
    \includegraphics[width=\textwidth]{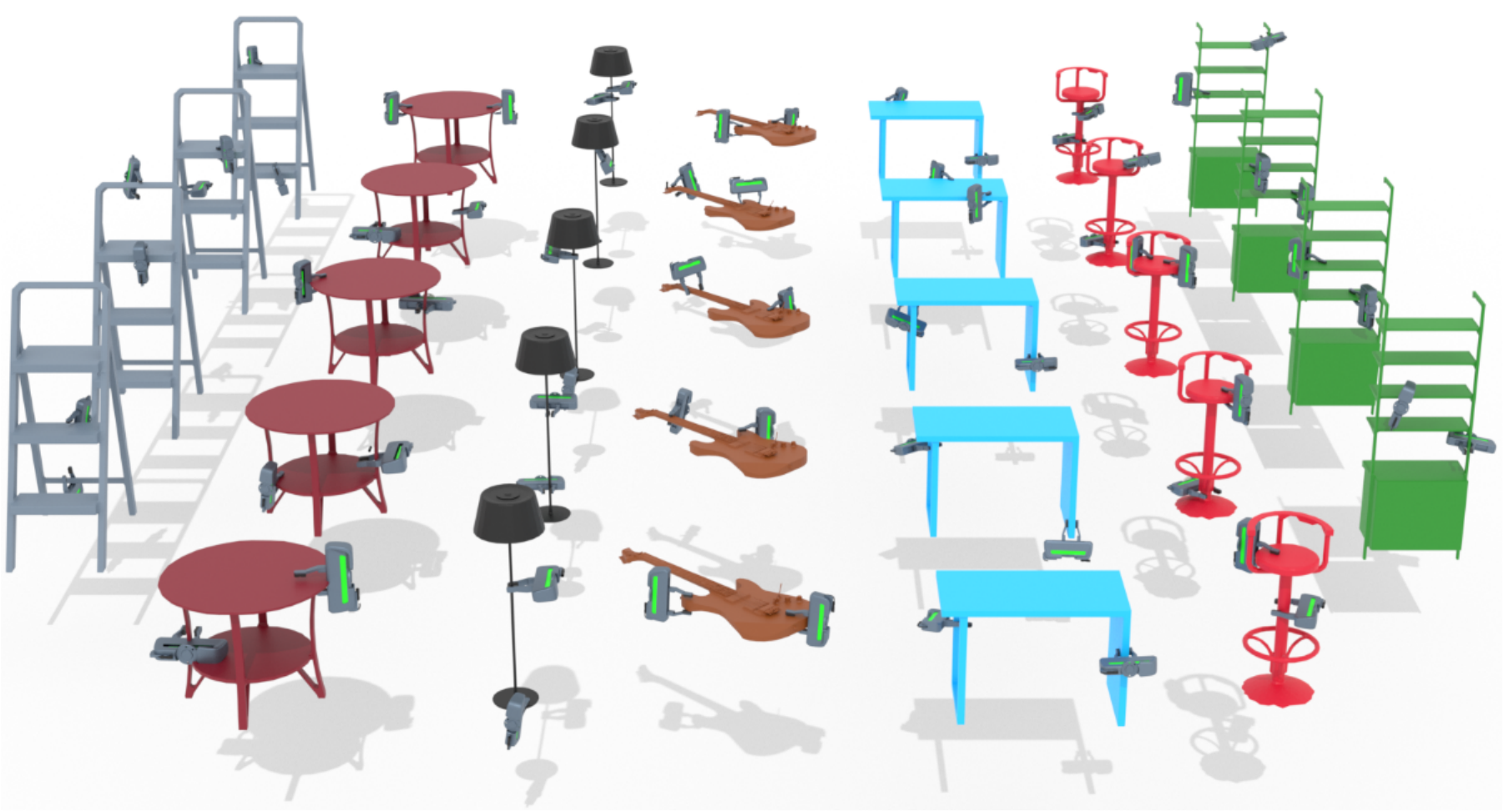}
    \caption{\textbf{Dataset Visualization:} We introduce DG16M, a large-scale dual-arm grasp dataset generated using optimization-based force-closure constraints to ensure stable and physically viable grasps. Our dataset provides high-quality grasp pairs, enabling better generalization for deep learning-based grasp generation models.}
    \label{fig:intro}
    \vspace{1mm} 
\end{center}
}]

\begin{abstract}

Dual-arm robotic grasping is crucial for handling large objects that require stable and coordinated manipulation. While single-arm grasping has been extensively studied, datasets tailored for dual-arm settings remain scarce. We introduce a large-scale dataset of 16 million dual-arm grasps, evaluated under improved force-closure constraints. Additionally, we develop a benchmark dataset containing 300 objects with approximately 30,000 grasps, evaluated in a physics simulation environment, providing a better grasp quality assessment for dual-arm grasp synthesis methods. Finally, we demonstrate the effectiveness of our dataset by training a Dual-Arm Grasp Classifier network that outperforms the state-of-the-art methods by 15\%, achieving higher grasp success rates and improved generalization across objects. Project page:
\url{https://dg16m.github.io/DG-16M/}

\end{abstract}

\input{Introduction}

\input{Related_Work}
\input{Methods}
\input{Experiments}
\input{Baselines}
\input{Conclusions}


\bibliographystyle{IEEEtran}

\bibliography{bibtex}

\end{document}

%% file: Introduction.tex
\section{Introduction}

Bimanual manipulation is essential in robotic applications involving handling large, heavy, or unwieldy objects that a single robotic arm cannot effectively grasp. Industries such as automation, logistics, assistive robotics, and assembly frequently require coordinated dual-arm strategies to ensure stable object handling~\cite{stabilize_to_act, trends_in_robot_manipulation, two_arms_better, cooperative_dualarm, dual_arm_survey}. 
A fundamental step in bimanual manipulation is establishing stable grasps on the object that distribute forces evenly, and minimize disturbances across various environments. 

Learning-based grasp planning methods have shown significant promise in the single-arm setting, leveraging large-scale datasets to improve generalization across diverse objects~\cite{contact_graspnet, 6dof_graspnet, cong, anygrasp, grasp_anything}. To sample such datasets, traditional analytical methods are used, that optimize grasp configurations based on object geometry and contact models, employing force-closure and wrench analysis to ensure grasp stability~\cite{dexnet2.0, graspit}. These approaches identify antipodal contact points, friction cones, and key stability metrics, such as the Ferrari-Canny epsilon metric~\cite{graspit}, to rank potential grasps.

\begin{table*}[t]
    \centering
    \renewcommand{\arraystretch}{1.3} 
    \setlength{\tabcolsep}{8pt} 
    \begingroup
    \normalsize 
    \begin{tabular}{l c c c c c}
        \hline
        \textbf{Dataset} & \textbf{Grasps per Obj} & \textbf{Grasp Label} & \textbf{Total Obj} & \textbf{Total Grasps} & \textbf{Gripper Type} \\  
        \hline
        ACRONYM~\cite{acronym_dataset} & 2000 & 6-DoF & 8872 & 17.7M & Single \\
        DA2~\cite{da2dataset} & up to 2001 & 6-DoF & 6327 & 9M & Dual \\
        \textbf{DG16M (ours)} & up to 4000 & 6-DoF & 4132 & 16M & Dual \\
        \hline
    \end{tabular}
    \endgroup
    \caption{Comparison of grasp datasets based on grasp density, labeling method, object count, total grasps, and gripper types supported.}
    \label{tab:grasp_datasets}
\end{table*}

Despite the availability of large-scale single-arm grasp datasets, they cannot be directly applied to dual-arm grasping. The primary challenge lies in effectively pairing single-arm grasps to create stable dual-arm grasp configurations~\cite{dual_arm_survey}. Unlike single-arm grasping, where stability is primarily determined by local contact properties, bimanual grasping requires considering additional constraints, such as object size, shape, and the coordination between both grippers. Extending grasp dataset generation to dual-arm settings introduces significant complexity due to these added constraints, making robust dataset construction and learning frameworks a challenging task. 
 
Existing dual-arm grasp dataset, DA2~\cite{da2dataset}, uses a simplified force-closure formulation that assumes constant normal forces at the contact points and lacks validation through physics simulation as shown in figure \ref{fig:qualitative_comparisons}. As a result, the dataset contains various types of failed grasps. In contrast, single-arm grasp datasets incorporate physics simulations to validate grasps, ensuring a higher degree of reliability~\cite{acronym_dataset}. Thus, there is a need for a high-quality dual-arm grasp dataset for the development of data-driven approaches for bimanual grasping.


To address this challenge, we introduce DG16M, a novel dataset that integrates a stricter force-closure formulation which analytically assesses grasp stability while accounting for varying external forces and gripper constraints. Our dataset consists of 4,143 objects of varying shapes and sizes from~\cite{da2dataset}, each with up to 4,000 dual-arm grasp pairs, resulting in a total of approximately 16 million grasps (a small subset is shown in Figure \ref{fig:intro}). Furthermore, we introduce a benchmark dataset containing 300 objects with approximately 30,000 grasps, verified through physics simulation-based evaluation using Isaac Gym~\cite{isaac_gym}.

To summarize our contributions:
\begin{itemize}
    \item We introduce DG16M, a large-scale dataset of 16 million dual-arm grasps, incorporating an improved force-closure validation to ensure physically stable and reliable grasp assessment. Through both classifier efficacy and analysis of grasp matrix, we demonstrate improvement in dual-arm grasp quality vis-\`{a}-vis prior datasets. 
    \item We develop an improved dual-arm grasp classifier leveraging neural descriptor fields, demonstrating that training on grasps vetted by our force-closure formulation yields higher success rates in simulation compared to prior methods. 
    \item We introduce a benchmark dataset of 300 objects with 30,000 simulation-verified grasps, providing a standardized evaluation for dual-arm grasping. By incorporating physics-based validation, our benchmark overcomes the limitations of force-closure analysis, ensuring a more practical and robust dual-arm grasp stability assessment.
    \item We also conduct ablation studies to demonstrate that sampling techniques like antipodal sampling alone cannot guarantee stable dual-arm grasp pairs. We show that these grasp pairs achieve stability only when further validated through our force-closure analysis.

\end{itemize}

%% file: Related_Work.tex
\section{Related Work}
\subsection{Grasp generation}

\begin{figure*}[!t]
    \centering
    \captionsetup{font=footnotesize}
    \includegraphics[width=\linewidth]{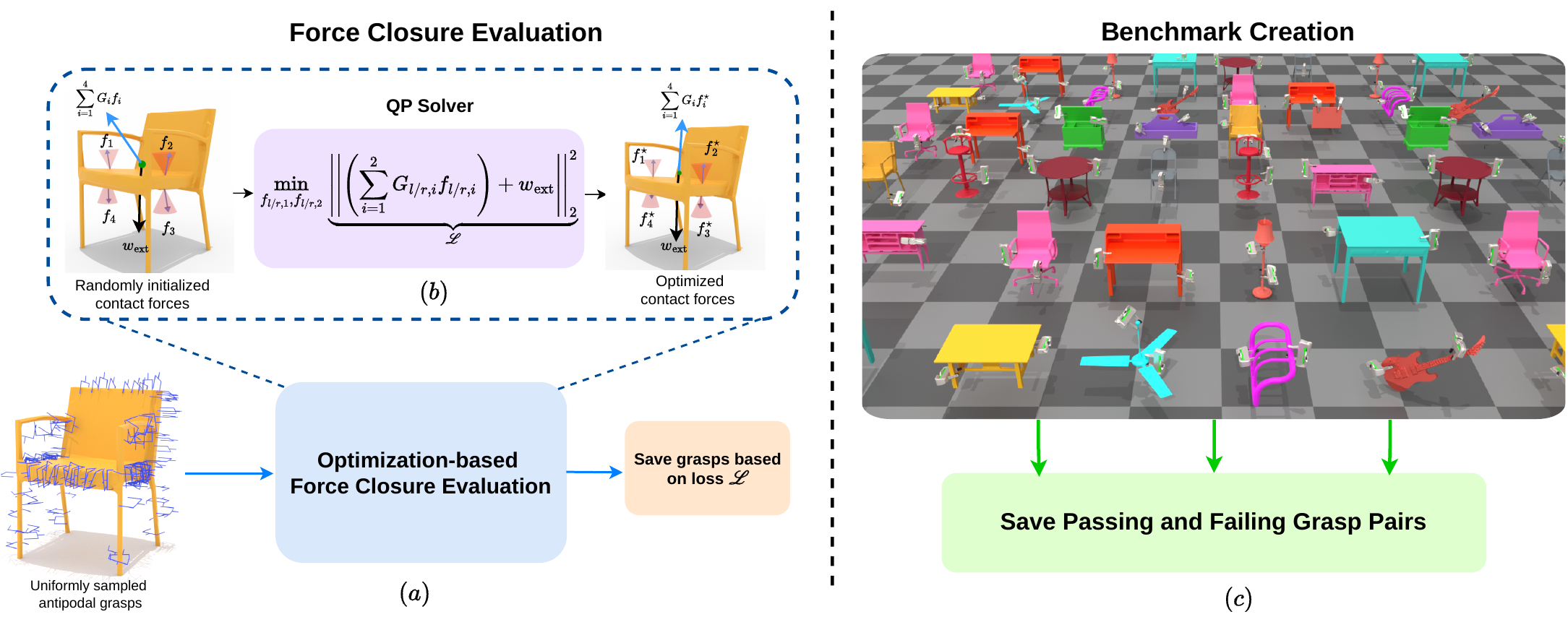}
    \caption{\textbf{Overview of the proposed method:}  
    \textbf{(a)} We start by sampling a large number of antipodal grasps on the object mesh, generate all possible grasp pairs, and apply distance-based pruning. These pairs are evaluated using an Optimizer-based Force Closure Evaluator, which checks if a valid set of contact forces can keep the object in equilibrium under an external wrench.  
    \textbf{(b)} The optimizer solves \eqref{eq:optimization_problem} under constraints \eqref{eq:constraint_1}, \eqref{eq:constraint_2}, enforcing gripper force limits and friction cone constraints. Valid grasp pairs are identified by thresholding the loss values, as shown in \eqref{eq:loss_tresholding}, ensuring that only stable grasps are retained while unstable ones are discarded.  
    \textbf{(c)} We select a subset of objects and evaluate their grasps in simulation to construct the benchmark dataset. This evaluation provides stability ground truth values for each dual-arm grasp, offering an objective assessment that is independent of predefined grasp quality metrics or assumptions.}

    \label{fig:pipeline}
\end{figure*}
Grasp synthesis approaches explore the space of possible grasp to identify stable and feasible grasps. One widely used approach is \textbf{approach-based sampling} \cite{sampling-approach1, sampling-approach2, sampling-approach3, sampling-approach4}, which aligns the gripper’s approach vector with the surface normal of the sampled object points. This alignment improves stability by ensuring perpendicular contact. However, it can introduce bias, potentially overlooking viable grasps that do not conform to the sampled surface normal~\cite{eppner2019billion}.

Another common strategy is \textbf{antipodal sampling} \cite{antipodal1}, which is extensively used in single-arm grasping datasets \cite{acronym_dataset, da2dataset, dexnet2.0, antipodal1}. This method selects opposing contact points that satisfy force-closure conditions, ensuring a stable grip, particularly for parallel-jaw grippers. Due to its simplicity and effectiveness, antipodal sampling has become a standard technique for generating stable grasps. For dual-arm grasping, we adopt the block antipodal sampling introduced in \cite{da2dataset}, as the first step in our grasp generation pipeline due to its ability to generate feasible contact points efficiently for large objects. Furthermore, we show in Section (\ref{enr}) that randomly pairing antipodal grasps is insufficient for stable dual-arm grasping, highlighting the necessity of force-closure validation.

\begin{figure*}[t]
    \centering
    \captionsetup{font=footnotesize}
    \includegraphics[width=\linewidth]{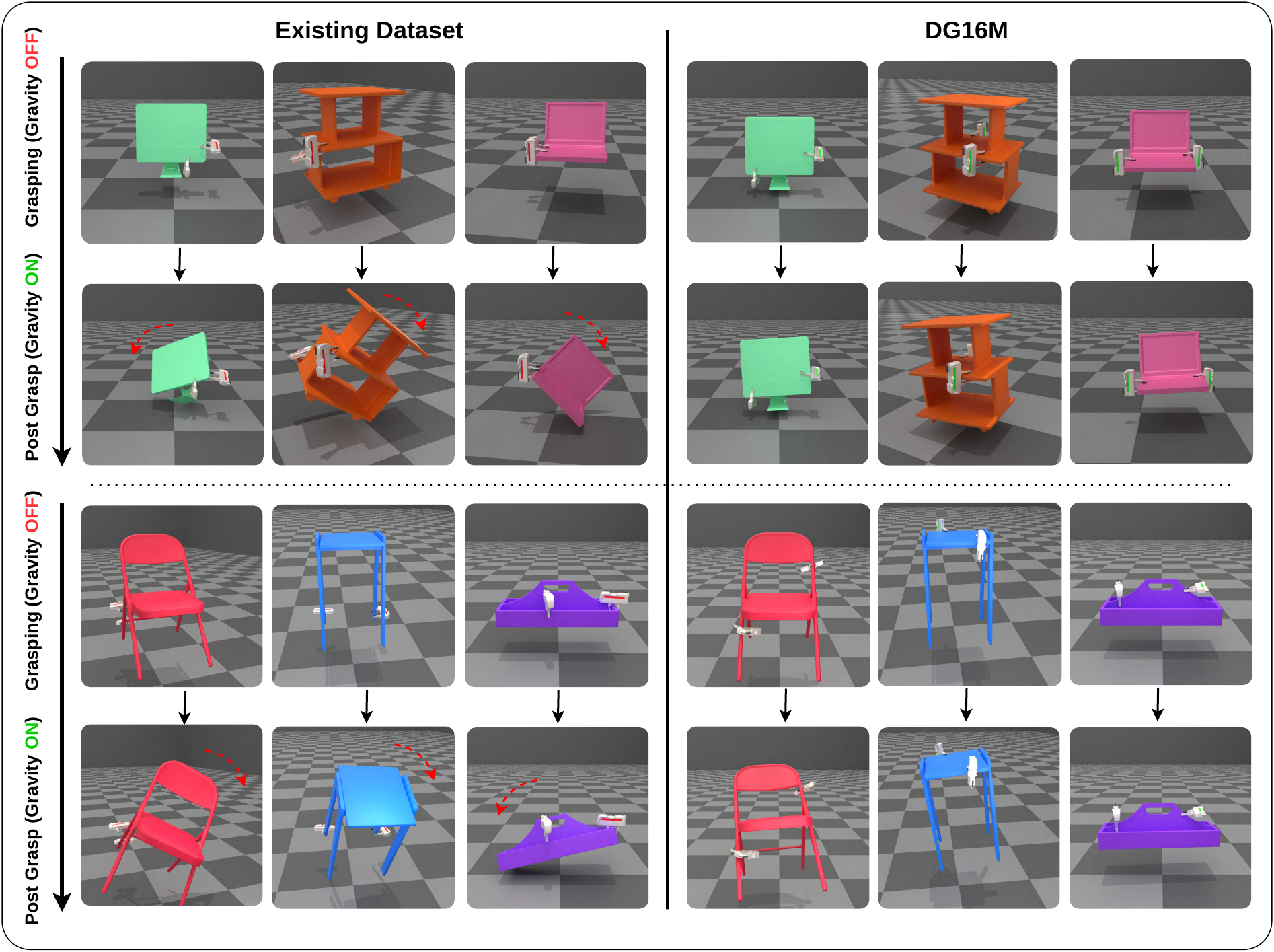}
    \caption{\textbf{Qualitative Comparison between the Existing DA2 Dataset and the Proposed DG16M Dataset}  
    We evaluate grasp stability by initializing the object and floating grippers at the grasp pose with gravity disabled. Once the grippers fully close, gravity is enabled. Grasps from the existing dataset often fail to hold the object in place (indicated by red dotted arrows) or lose contact due to instability, which arises from issues such as grasp pairs being too close, positioned on the same side of the center of mass, or lacking force balance. In contrast, our dataset provides more stable and physically viable grasps, ensuring successful object retention without failures. Floating grippers are shown in white. \textit{(Zoom in for a clearer view)}}
    \label{fig:qualitative_comparisons}
\end{figure*}

\subsection{Force Closure}
Force closure is a fundamental property in robotic grasping, ensuring that an object can be securely held against external disturbances. A grasp achieves force closure if the set of contact forces can counteract any external wrench acting on the object. Traditionally, force closure is computed by constructing a convex hull over discretized friction cones in wrench space under the assumption that the total contact forces sum to one \cite{Murray1994AMI, fc1}.
\cite{diff_fc} proposed two relaxations to the traditional formulation, namely, zero friction and equal magnitude of contact forces, to make the formulation differentiable. Even though this formulation is useful for a human hand, the assumptions made pose certain challenges when adapting it to a dual-gripper setting~\cite{da2dataset}.
PhyGrasp~\cite{phygrasp} uses an optimization-based force closure formulation to solve for the contact forces. They sample contact points uniformly across the object without any constraints in order to find an affordance map, resulting in some of these grasps being practically infeasible.
We propose an optimization-based constrained force closure formulation that addresses the issues with using a relaxed~\cite{diff_fc} or unconstrained~\cite{phygrasp} formulation for a dual-gripper setting.



\subsection{Benchmarking Datasets for Grasping}
In the field of robotic grasping, the development of benchmarking datasets has been pivotal for advancing grasp synthesis and manipulation techniques, with most efforts focused on \textbf{single-arm} grasping. Several large-scale single-arm datasets, such as \cite{Graspnet1B, Cornell, Jacquard, dexnet2.0, regrard}, provide RGB-D images of objects along with annotated grasps, either in the form of 6D grasp poses or bounding box markings. In addition, datasets like \cite{cong, acronym_dataset} focus on synthetic point clouds and grasp annotations, expanding the range of objects available for training. While these datasets are extensive and contain a large number of grasps, they are primarily designed for small-sized objects, partly due to the physical constraints of single-arm manipulation. Therefore, they typically use object meshes from YCB \cite{ycb} and GSO \cite{gso} or specific categories of ShapeNet \cite{shapenet}(e.g. bottle, mug, cup, etc.)


Existing dual-arm grasping datasets are limited, with DA2 \cite{da2dataset} being the first large-scale dual-arm grasp dataset, containing approximately 9 million grasps across a wide range of large objects selected from the ShapeNet dataset~\cite{shapenet} (refer to Table \ref{tab:grasp_datasets}). While this dataset is extensive, it follows a relatively relaxed set of force closure constraints that do not account for actuator constraints, increasing the gap between the grasps in the dataset and the grasps which are practically feasible.
Conversely, DG16M employs an optimization-based analytical force closure formulation that can account for the actuator constraints and provide optimal contact forces that counteract external wrenches.
This results in well-separated positive and negative grasp pairs, enabling classifiers like \cite{cgdf} and \cite{pointnetgpd} to learn effectively. 
Our dataset can readily be integrated with frameworks such as our previous work DA-VIL \cite{davil} to achieve better inferences.

%% file: Methods.tex
\section{Dataset Generation}
\label{section:dataset_generation}

In this section, we describe the pipeline used to generate grasps for our DG16M dataset as shown in Figure \ref{fig:pipeline}.
The pipeline consists of two key steps: (1) generating grasp candidates through antipodal sampling, (2) evaluating grasp quality using an optimization-based force closure formulation. 

\subsection{Grasp Sampling}
\label{method:grasp_sampling}

To generate grasp candidates, we employ \textbf{antipodal grasp sampling}, a method commonly used for grasp pose generation \cite{dexnet2.0}. The process begins by randomly selecting a point on the object's surface along with its corresponding normal. A second contact point is then found by tracing a ray within a constrained region around the normal direction, ensuring that the two contact points satisfy antipodal conditions. The allowable range for this ray is determined by a threshold $\gamma = \tan(\alpha/2)$, where $\alpha$ is the cone angle given by $\alpha = \tan^{-1}(\mu)$, with $\mu$ representing the friction coefficient. After identifying both contact points, we verify that the grasp configuration does not result in any collision between the grippers and the object. Any grasp that leads to penetration or overlaps with the object is discarded, ensuring that the final sampled grasps are physically valid and free of collisions. We initially sample 500 single-arm grasps per object, which are then combined exhaustively to generate all possible unique combinations of dual-arm grasp pairs. To improve diversity and prevent redundancy, we apply a distance-based pruning step to remove grasp pairs that are too close to each other.  Following this filtering process, we are left with an average of 30,000 to 80,000 dual-arm grasp pairs per object. These candidate grasp pairs are then passed through our Optimizer-based Force Closure framework to ensure that only physically stable and feasible grasps are retained.

\subsection{Optimization-based Force Closure Formulation}

The quality of the generated set of candidate grasp pairs is then evaluated by solving the force closure optimization problem. The goal of this optimization is to determine the contact forces that can resist an external wrench ($\boldsymbol{w}_\text{ext}$) acting on the object at the centre of mass in the object frame,  while satisfying the friction cone constraints and thus obeying force closure.

The dual-arm setup equipped with two dedicated two-fingered rigid grippers, results in four contact points on the object's surface while grasping. Let the position and orientation of the contact frame relative to the object frame be represented as $\boldsymbol{p}_{l/r,i} \in \mathbb{R}^3$ and $\boldsymbol{R}_{l/r,i} \in \mathbb{SO}(3)$, where $l/r$ represents the left/right grasp and $i$ denotes the $i^{th}$ contact point of the respective gripper, with $i=\{1,2\}$. 

The force applied by the $i^{th}$ contact point on the object in the contact frame is given by $\boldsymbol{f}_{l/r,i} = \boldsymbol{f}_{t_{l/r,i}} + \boldsymbol{f}_{n_{l/r,i}}$, where $\boldsymbol{f}_{t_{l/r,i}}$ and $\boldsymbol{f}_{n_{l/r,i}}$ represent the tangential and normal components of the contact forces, respectively. The grasp matrix \( \boldsymbol{G}_{l/r,i} \) maps the contact force $\boldsymbol{f}_{l/r,i}$ to the object frame and is given by 
\[
\boldsymbol{G}_{l/r,i} = \begin{bmatrix}
\boldsymbol{R}_{l/r,i} & \boldsymbol{0} \\
[\boldsymbol{p}_{l/r,i}]_{\times} \boldsymbol{R}_{l/r,i} & \boldsymbol{R}_{l/r,i}
\end{bmatrix} \cdot \boldsymbol{B},
\]
where $\boldsymbol{B} = [\mathbf{I}_{3 \times 3} \ \mathbf{0}_{3 \times 3}]^T$ represents the contact basis for the Point Contact With Friction (PCWF) model \cite{Murray1994AMI} and \([\boldsymbol{p}_{l/r,i}]_{\times} \in \mathbb{R}^{3\times3} \) is the skew-symmetric matrix of the contact position \( \boldsymbol{p}_{l/r,i} \).

Candidate grasp pairs are validated using a rank test to ensure grasp feasibility. The grasp matrix \( \boldsymbol{G} \) is formed by concatenating individual contact grasp matrices. Grasp pairs whose grasp matrix \( \boldsymbol{G} \) is not full rank are considered infeasible, indicating insufficient wrench controllability.
 

The remaining grasp pairs proceed to the next stage, where the optimal force distribution is determined by formulating the force closure optimization problem as a Second-Order Cone Program (SOCP). The objective function is:

The objective function is:
\begin{equation}
    \mathcal{L} = \bigg \| \bigg (\sum_{i=1}^{2} \boldsymbol{G}_{l/r,i} \cdot \boldsymbol{f}_{l/r,i} \bigg)\ + \boldsymbol{w_{\text{ext}}} \bigg \|^{2}_{2}
\end{equation}

The optimization problem is formulated as:
\begin{equation}
    \min_{\boldsymbol{f}_{l,1}, \boldsymbol{f}_{l,2}, \boldsymbol{f}_{r,1}, \boldsymbol{f}_{r,2}} \mathcal{L},
    \label{eq:optimization_problem}
\end{equation}
subject to,

\begin{equation}
     \| \boldsymbol{f}_{l/r,i} \|_{2} \leq \boldsymbol{f}_{\text{high}}, \quad \forall i \in \{1,2\}
     \label{eq:constraint_1}
\end{equation}
\begin{equation}
    \| \boldsymbol{f}_{t_{l/r,i}} \|_{2} \leq \mu \boldsymbol{f}_{n_{l/r,i}}, \quad \forall i \in \{1,2\}
    \label{eq:constraint_2}
\end{equation}

Constraint \eqref{eq:constraint_1} ensures that the contact forces are within the physical limits of the grippers (where \(\boldsymbol{f}_\text{high}\) is the maximum contact force the gripper can exert) while constraint \eqref{eq:constraint_2} ensures that the contact force lies within the friction cone to prevent slipping. 

\subsection{Dual-arm Grasps Selection}
The optimization problem is solved using a convex optimization solver, CVXPY \cite{cvxpy}. The quality of the grasp is evaluated based on the optimal value of the objective function, which indicates how well the grasp can resist the external wrench. We define a grasp to be successful if the optimization loss satisfies, 
\begin{equation}
    \mathcal{L} \leq 10^{-5}
    \label{eq:loss_tresholding}
\end{equation}
This threshold ensures that the net wrench on the object is close to zero, indicating a stable grasp configuration. Grasps failing to meet this criterion are discarded to maintain dataset quality. Finally, to construct a reliable and diverse dataset, we retain successful grasp pairs that satisfy both the rank condition and the force closure condition per object. This selection process ensures that only high-quality, physically feasible grasps are included, making the dataset well-suited for benchmarking dual-arm grasping algorithms.

The force closure formulation in the existing dataset differs from our approach in several key aspects. DA2 does not compute optimal force values explicitly, focusing instead on generating grasp candidates through antipodal sampling and evaluating their stability through force closure. In contrast, our formulation directly incorporates the computation of optimal contact forces, ensuring that the grasp can resist external disturbances while satisfying friction cone constraints. Furthermore, our approach explicitly accounts for the physical limits of gripper forces, making it more practical for real-world applications where gripper capabilities play a critical role. By incorporating these constraints, we ensure that the generated grasps are not only theoretically stable but also physically feasible, allowing for reliable execution within the hardware’s operational limits.The superior performance of classifiers trained on our dataset in physics simulations further validates the effectiveness of our dataset in enabling robust and stable dual-arm grasps.

%% file: Experiments.tex
\section{Benchmark Dataset Creation}
\label{section:bechmark}
Benchmarking grasping performance is challenging due to variations in controllers, grippers, and robotic arms. While it is practically infeasible to be agnostic to gripper designs, we ensure that a successful grasp renders the object immovable under external forces. To establish this benchmark, we define the essential criteria for grasp stability by ensuring that the grasp prevents object displacement when subjected to external disturbances. Notably, grasps that appear successful based on contact points alone do not always translate into stable grasps, as they often involve area contact rather than point contact.

To construct our benchmark dataset, we select 300 objects from our object list, designating them as part of the unseen object split. This subset serves as the test set for subsequent experiments, allowing us to evaluate grasp stability.

We employ the grasp generation technique outlined in Section~\ref{section:dataset_generation} to generate valid dual-arm grasps for these objects. These grasps are then evaluated in the Issac Gym simulator \cite{isaac_gym} using physics-based validation. During evaluation, the Franka Panda grippers are initialized at the designated grasp pose with gravity disabled to prevent premature object displacement. Each gripper can exert a maximum gripping force of 70N \footnote{\href{https://download.franka.de/documents/100010_Product Manual Franka Emika Robot_10.21_EN.pdf}{Franka Emika Robot's Instruction Handbook}}. Once the grippers fully close, gravity is enabled, and forces are applied to stabilize the object. A grasp is deemed successful if the object remains stationary under these conditions. Otherwise, it is classified as a failure.


This evaluation process establishes a ground truth dataset, independent of predefined grasp quality metrics or assumptions, providing an objective and practical measure of grasp success. As a result, we compile a benchmark dataset of 30,000 dual-arm grasps, including their associated objects and simulation-verified ground truth labels.

\section{Experiments and Results} 
\label{enr}
In this section, we present the evaluation metrics for dual-arm grasps and analyze the performance of our proposed baselines on the DG16M and DA2 datasets.



\textbf{Metrics:} We evaluate grasps using two complementary approaches: \textbf{Force Closure Evaluation (FCE)} and \textbf{Grasp Success Rate (GSR)}. These metrics measure the feasibility and robustness of grasps under static and dynamic conditions, respectively. 

\begin{table}[t]
    \centering
    \renewcommand{\arraystretch}{1.4} 
    \setlength{\tabcolsep}{2pt} 
    \begingroup
    \normalsize 
    \begin{tabular}{l  cc  cc}
        \hline
        \multirow{2}{*}{\textbf{Model}} & \multicolumn{2}{c}{\textbf{Our Dataset}} & \multicolumn{2}{c}{\textbf{DA$2$ Dataset}} \\  
        \cline{2-5}  
        & FCE(\%)$\uparrow$ & GSR(\%)$\uparrow$ & FCE(\%)$\uparrow$ & GSR(\%)$\uparrow$ \\  
        \hline
        \addlinespace[1.5mm]
        \shortstack{Dual- \\ PointNetGPD} & 69.5 & 61.22 & 61.8 & 53.26\\   
        \textbf{CGDF-Classifier} & \textbf{88.73} & \textbf{76.34} & \textbf{65.25} & \textbf{54.33} \\  
        \hline
    \end{tabular}
    \endgroup
    \caption{Performance comparison of different models on our dataset and DA2 dataset.}
    \label{tab:results}
\end{table}

\textit{Force Closure Evaluation (FCE)}:
This evaluation is an analytical method that assesses grasp feasibility through force closure analysis. The optimizer computes the required contact forces at the grasp points, ensuring static equilibrium while satisfying friction constraints. A grasp is considered successful if these forces can counteract external disturbances and maintain the object's stability.

\textit{Grasp Success Rate (GSR)}:
This evaluation method simulates real-world grasp execution using floating grippers in Isaac Gym Simulator \cite{isaac_gym}, as described in Section \ref{section:bechmark}. It measures not only the theoretical feasibility of grasps but also their practical stability, ensuring they are viable for real-world execution.  


\textbf{Baseline methods:}
To assess the efficacy of our dual-arm grasping framework, we benchmark it against two baseline grasp classification methods: (1) We adapt PointNetGPD~\cite{pointnetgpd} to a dual-arm setting as done in~\cite{da2dataset}. (2) We use SE(3)-aware grasp diffusion models~\cite{cgdf, se3dif} to give us grasp descriptors and train a classifier head on the features. We call these two baselines Dual-PointNetGPD, and CGDF-Classifier. 

We use the Dual-PointNetGPD architecture, as described in \cite{da2dataset}. To adapt the original PointNetGPD architecture for evaluating dual-arm grasps, firstly, the gripper templates are transformed to their grasp poses using transformation matrix \( \boldsymbol{H} \in SE(3) \). To obtain grasp region information, the nearest 512 points from the object point cloud $P_0$ are selected around each transformed gripper pose, forming two local point cloud regions, $P_1, P_2$. The combined set of sampled points, \( P = P_1 \cup P_2 \), serves as a local grasp representation, which is then passed through the PointNetGPD architecture to extract a feature vector. This feature vector is subsequently processed by a Multi-Layer Perceptron (MLP) to classify whether the grasp is successful or not.


Recent approaches have explored incorporating both object geometry and grasp descriptors into the learning process. For the CGDF-Classifier, we build upon \cite{cgdf} by integrating a classification head following the vision encoder and feature encoder outputs, which is based on neural descriptor fields \cite{ndf}. The network receives the object point cloud \( P_o \in \mathbb{R}^{M \times 3} \), consisting of $M$ points, and the gripper template point cloud \( P_g \in \mathbb{R}^{N \times 3} \), consisting of $N$ points. The gripper templates are transformed into the object’s plane using the transformation matrix  \( \boldsymbol{H} \in SE(3) \). This design enables the model to learn both object features and the relative grasp information within the object’s reference frame. Based on these feature vectors, the classifier determines whether a given grasp is successful or not.

\textbf{Results:} As shown in Table ~\ref{tab:results}, models trained on our dataset consistently outperform those trained on DA2 across both the metrics. The higher performance of both the models on our dataset highlights the superior quality of our dataset, which employs well-defined force closure formulation. This results in well separated positive and negative grasp pairs, enabling models to learn more effectively.

\begin{table}[t]
    \centering
    \renewcommand{\arraystretch}{1.25} 
    \setlength{\tabcolsep}{9pt} 
    \normalsize 
    \begin{tabular}{l  cc}
        \hline
        \multirow{2}{*}{\textbf{Generation Technique}} & \multicolumn{2}{c}{\textbf{Evaluation Metrics}} \\
        \cline{2-3}  
        & FCE(\%) $\uparrow$ & GSR(\%) $\uparrow$ \\  
        \hline
        Random Dual-Arm Pairs & 16.33 & 23.66 \\  
        Farthest-Grasp Pairs & 26.16 & 38.45 \\  
        \textbf{DG16M (Ours)} & \textbf{100.0} & \textbf{76.33} \\
        \hline
    \end{tabular}
    \caption{Ablation study comparing different dual-arm grasp generation techniques.}
    \label{tab:ablation}
\end{table}


A key observation is that models trained on DA2 exhibit near-random classification performance in simulation, as indicated by GSR values hovering around 50\%. This suggests that DA2-trained models struggle to reliably distinguish between successful and failed grasps, classifying them almost at random. Although DA2 incorporates a structured grasp stability criterion, its grasp distributions may not transfer effectively to simulation-based evaluations, potentially limiting generalization. In contrast, our dataset provides a better grasp distribution, ensuring that models learn from more stable and physically meaningful examples. Unlike DA2-trained models, our models not only achieve high theoretical grasp feasibility (FCE) but also demonstrate strong performance in realistic physics simulations (GSR), proving their viability beyond just theoretical metrics.

Additionally, we observe a slight performance gap between FCE and GSR, largely due to the challenges of translating theoretical grasp stability into practical execution in a simulated environment. Factors such as unmodeled physical effects, contact dynamics, and complex gripper-object interactions contribute to this discrepancy.


\begin{figure}[t]
    \centering
    \captionsetup{font=footnotesize}
    \includegraphics[width=\linewidth]{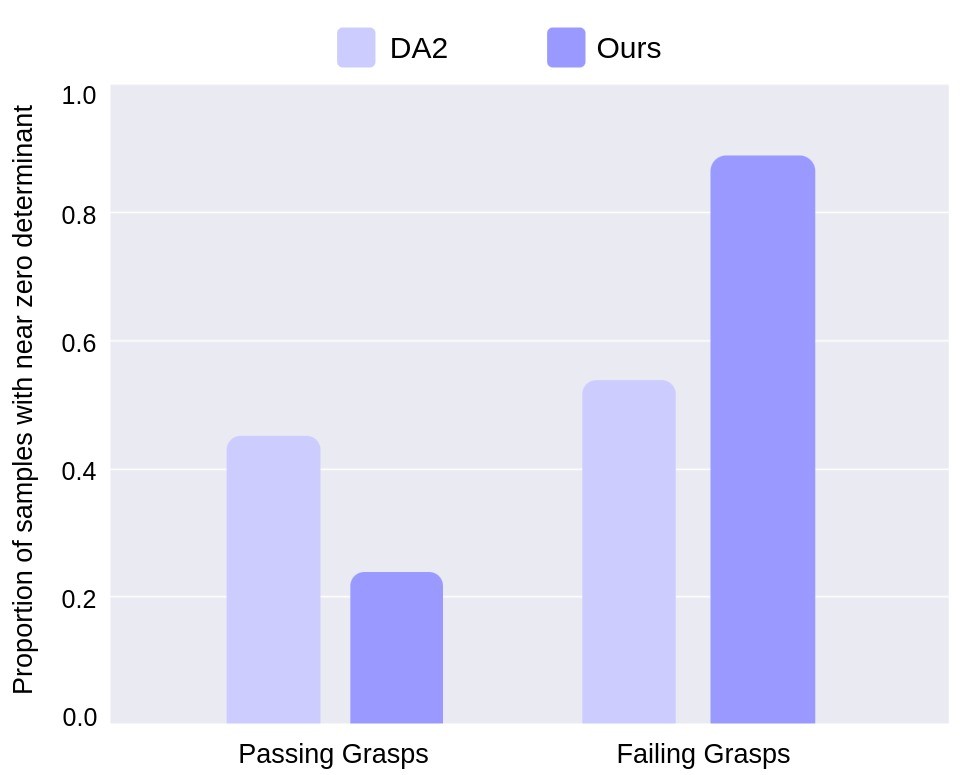}

    \caption{\textbf{Grasp Stability Analysis}. We compute grasp stability using \( Q(\boldsymbol{G}) = \sqrt {\det({\boldsymbol{G}\boldsymbol{G}^T)}}\), which measures the volume of the grasp wrench space. Our dataset follows the expected trend of having large $Q(\boldsymbol{G})$ values for passing grasps while unstable grasps having near-zero $Q(\boldsymbol{G})$ values. However, DA2 shows both stable and unstable grasps with similar low $Q(\boldsymbol{G})$ values, indicating less manipulable grasps.}
    \label{fig:plot_determinant}
\end{figure}

Further, we validate the quality of our dataset by analyzing grasp stability, $Q(\boldsymbol{G})$ = \(\sqrt {\det({\boldsymbol{G}\boldsymbol{G}^T)}}\), where $\boldsymbol{G}$ is the grasp matrix, which quantifies the volume of the grasp wrench space, reflecting the grasp's ability to transfer contact forces into wrenches that stabilize the object \cite{GG}. A large $Q(\boldsymbol{G})$ indicates a robust grasp with good manipulability, while a near-zero $Q(\boldsymbol{G})$ suggests a failing grasp that is close to a singular configuration and unable to resist external disturbances. As shown in Figure \ref{fig:plot_determinant}, our dataset (purple bars) exhibits a clear separation between successful and failed grasps, with most failures and very few successes having near-zero determinants. In contrast, the DA2 dataset (pink bars) shows little distinction between passed and failed grasps, as its determinant values do not follow the expected trend.
Many successful and failed grasps alike exhibit near-zero determinants, undermining their reliability as grasp stability indicators.

\textbf{\textit{Qualitative Results.}} Figure \ref{fig:qualitative_comparisons} provides a qualitative comparision of grasps generated by models trained on our dataset versus DA2. For large objects with complex geometries, models trained on DA2 often detect grasps that fail to satisfy force closure or dynamic stability conditions. In contrast models trained on our dataset detect grasps that are both theoretically feasible and practically executable. DA2-trained models often produce unsuccessful grasps because the dataset's grasp samples frequently fail to secure the object or maintain contact. This instability arises from factors such as grasp pairs being too close together, positioned on the same side of the center of mass, or lacking proper force balance, all of which compromise grasp stability. These failures are primarily due to the relaxed force closure condition used in DA2, which does not account for optimal contact forces and physical limits of the grippers, leading to infeasible grasps.

\textbf{Ablations.} Our key design decision in generating dual-arm grasps is the use of an optimizer-based force closure formulation to select stable grasp pairs from candidate samples. To evaluate the necessity and effectiveness of this approach, we conduct ablation studies comparing our grasp selection method against two alternative strategies. (1) Random Pairing of Single-Arm Grasps, where single-arm grasps are sampled and randomly combined to form paired grasps, and  
(2) Diametrically Opposite Pairing, where single-arm grasps are sampled, and the farthest grasps are paired together.
The results are presented in Table \ref{tab:ablation}.

\begin{enumerate}
    \item \textbf{\textit{Random Pairing of Single-Arm Grasps}:} After sampling grasps through antipodal sampling, random pairing is performed. This naive approach results in a low FCE of 16.33\% and GSR of 23.66\%, indicating that most grasp pairs fail to maintain stability. Since the pairing is done arbitrarily, the resulting grasps often lack proper force balance and fail to satisfy stability constraints, leading to frequent failures in both theoretical evaluation and simulation tests. This ablation highlights why merely relying on antipodal sampling is insufficient and demonstrates the necessity of incorporating force closure to ensure stable dual-arm grasps.
    \item \textbf{\textit{Pairing Diametrically Opposite Grasps}:} Ensuring that grasp pairs are geometrically opposite improves FCE to 26.16\% and GSR to 38.45\%. While this method guarantees spatial separation, it incorrectly assumes that geometric opposition alone ensures stability. In reality, it ignores force and torque equilibrium, causing many grasp pairs to still fail under external disturbances.
    \item \textbf{\textit{Force Closure-Based Pairing}:} Our method, which explicitly enforces force closure constraints, achieves 100\% FCE and a GSR of 76.33\%, significantly outperforming both alternative approaches. By ensuring that grasp pairs satisfy both spatial separation and physical stability, our approach produces grasps that are theoretically stable and also successful in simulated environments.
\end{enumerate}

The ablation studies validate our force closure approach. Random pairing lacks physical reasoning, while diametrically opposite pairing improves but remains limited. Force closure, by incorporating physical constraints, delivers the best performance, underscoring its importance in dual-arm grasp generation.

%% file: Conclusions.tex
\section{Conclusions and Future Work}


We introduced DG16M, a large-scale dual-arm grasp dataset with 16 million grasps, addressing limitations in existing datasets like DA2. DG16M enforces force-closure constraints and incorporates simulation-based validation to ensure both theoretical and physical grasp stability. Additionally, we provide a benchmark of 30,000 simulation-verified grasps across 300 objects for robust evaluation.

Deep learning-based grasp classifiers trained on DG16M, including Dual-PointNetGPD and CGDF-Classifier, achieved significantly higher Grasp Success Rates (GSR) and Force Closure Evaluation (FCE) scores than those trained on DA2. This underscores the importance of structured, high-quality grasp distributions in improving generalization across diverse objects and configurations, making DG16M a strong foundation for dual-arm grasping research.

Future work could explore real-world grasp trials with DG16M to validate its effectiveness beyond simulation. Additionally, it could investigate adaptive grasping strategies that dynamically adjust parameters based on object properties and external disturbances, enhancing the versatility and reliability of dual-arm robotic systems in real-world tasks.